\documentclass[10pt,twocolumn,letterpaper]{article}

\usepackage{iccv}
\usepackage{times}
\usepackage{epsfig}
\usepackage{graphicx}
\usepackage{amsmath}
\usepackage{amssymb}

\usepackage{multirow}
\usepackage{enumitem}
\usepackage{wrapfig}
\usepackage[cjk]{kotex}
\usepackage{xcolor,colortbl}
\usepackage{booktabs}
\usepackage{amsfonts}       %
\usepackage{microtype}      %
\usepackage{xspace}
\usepackage{amsthm}
\usepackage[font=footnotesize,skip=3pt,subrefformat=parens]{subcaption}
\usepackage[font=small]{caption}
\usepackage{multirow}
\usepackage[accsupp]{axessibility}

\usepackage[pagebackref=true,breaklinks=true,letterpaper=true,colorlinks,bookmarks=false, citecolor=teal]{hyperref}

\iccvfinalcopy %

\def\httilde{\mbox{\tt\raisebox{-.5ex}{\symbol{126}}}}

\ificcvfinal\fi

\newcommand{\rulesep}{\unskip\ \vrule height 34mm}

\newcommand\ours{\texttt{GA}\xspace}

\renewcommand\footnotemark{}

\begin{document}

\title{Gramian Attention Heads are Strong yet Efficient Vision Learners}
\author{Jongbin Ryu$^\star$\thanks{$^\star$Equal contribution.}\\
Ajou University\\
{\tt\small jongbinryu@ajou.ac.kr}
\and
Dongyoon Han$^\star$\\
NAVER AI Lab\\
{\tt\small dongyoon.han@navercorp.com}
\and
Jongwoo Lim$^\dagger$\thanks{$^\dagger$This work was done when Jongwoo Lim was at Hanyang University.}\\
Seoul National University\\
{\tt\small jongwoo.lim@gmail.com}
}

\maketitle
\ificcvfinal\thispagestyle{empty}\fi

\begin{abstract}
We introduce a novel architecture design that enhances expressiveness by incorporating multiple head classifiers (\ie, classification heads) instead of relying on channel expansion or additional building blocks.
Our approach employs attention-based aggregation, utilizing pairwise feature similarity to enhance multiple lightweight heads with minimal resource overhead. We compute the Gramian matrices to reinforce class tokens in an attention layer for each head. This enables the heads to learn more discriminative representations, enhancing their aggregation capabilities. Furthermore, we propose a learning algorithm that encourages heads to complement each other by reducing correlation for aggregation. Our models eventually surpass state-of-the-art CNNs and ViTs regarding the accuracy-throughput trade-off on ImageNet-1K and deliver remarkable performance across various downstream tasks, such as COCO object instance segmentation, ADE20k semantic segmentation, and fine-grained visual classification datasets. The effectiveness of our framework is substantiated by practical experimental results and further underpinned by generalization error bound. We release the code publicly at: \href{https://github.com/Lab-LVM/imagenet-models}{https://github.com/Lab-LVM/imagenet-models}.
\end{abstract}

\section{Introduction}
\label{sec:intro}
Supervised learning opened the door to the emergence of a plethora of milestone networks~\cite{vgg,resnet,efficientnet,convnext,vit,swin} that achieved significant success on ImageNet~\cite{imagenet}. Training a single network with the cross-entropy loss has been a simple standard for image classification; this also holds for training multiple networks or multiple features~\cite{newell2016stacked,fpn,yu2018deep_dla,sun2018fishnet,dft_eccv_2018,du2020spinenet}. The methods of extracting multiple features at different stages aim to aggregate diversified features from an architectural perspective. %
Previous works~\cite{newell2016stacked,fpn,sun2018fishnet,du2020spinenet} expand their architectures by incorporating many trainable layers to refine features, relying on the architectural perspective. Their success is likely attributed to extra heavy layers that promote feature diversification. However, it remains uncertain whether the architectures effectively promote learning favorable less-correlated representations~\cite{Dropout,cogswell2015reducingdecorr, GCFN,huang2018decorrelated, hua2021featuredecorr}. Additionally, their intentional design for high network capacity with numerous trainable parameters increases computational demands.

In this paper, we present a new design concept of deep neural networks that learns multiple less-correlated features at the same time. Since motivated by feature aggregation methods~\cite{newell2016stacked,fpn,yu2018deep_dla}, we aim to avoid excessively over-parameterized networks and realize performance improvement through the learning of multiple less-correlated features. Our architecture consists of multiple shallow head classifiers on top of the backbone instead of increasing depth or width and without employing complicated decoder-like architectures. Therefore, it is evident that our architecture offers a speed advantage, but the potentially limited expressiveness with lightweight heads is problematic. A question that naturally arises is \textit{how can we improve the network capacity of shallow heads with limited trainable parameters?}

Our answer centers on the idea of introducing the Gramian matrix~\cite{gatys2015neural,yim2017gift} combined with the attention module~\cite{vaswani2017attention}. The Gramian is identical to the bilinear pooling~\cite{bilinear_pooling} that collects the feature correlations so that the attention can further leverage the information of the Gramian of features. Specifically, we compute the Gramian matrices of each output of heads before the final predictions and feed them into each attention as the query, which brings the pairwise similarity of features. This design principle is naturally scalable to any backbones, including Convolutional Neural Networks (CNNs)~\cite{resnet,mobilenetv3, efficientnet, regnet,convnext}, Vision Transformers (ViTs)~\cite{vit,swin}, and hybrid architectures~\cite{coatnet,pflayer,poolformer,tu2022maxvit}.

We further introduce a learning algorithm that forces each head to learn different and less correlated representations. The algorithm is based on the proposed decorrelation loss, which performs like an inverse knowledge distillation loss~\cite{kd_hinton}. Our proposed framework compels lightweight heads to learn distinguished and enhanced representations. It turns out that our trained models can replace complicated ones through empirical evaluations and effectively be generalized to other downstream tasks.

We evaluate our models by training them on the CIFAR100~\cite{cifar} and ImageNet-1K~\cite{imagenet} datasets to showcase the effectiveness of our proposed method. We systematically compare the competing models with similar computational budgets, including the throughput of a model; ours beat the state-of-the-art CNNs, ViTs, and hybrid architectures in the accuracy and throughput trade-off. 
We also show the transferability of our pretrained models to downstream tasks, including instance segmentation and semantic segmentation on COCO~\cite{coco2017} and ADE20k~\cite{ade20k}, respectively, and fine-grained visual classification datasets\footnote{Experimental results of the fine-grained visual recognition tasks are found in Table~\ref{tbl:mem}.}, including CUB-200~\cite{dataset_CUB}, Food-101~\cite{food101}, Stanford Cars~\cite{stanford_cars}, FGVC Aircraft~\cite{fgvc_aircraft}, and Oxford Flowers-102~\cite{flower102}. 

Finally, we analyze the efficacy of our design elements, including hyper-parameters via Strength and Correlation theory~\cite{GCFN, random_forest}. The theory manifests as Correlation is lowered while Strength increases, and classifiers learn a highly generalizable representation. While Correlation and Strength are usually proportional, our framework has elements lowering Correlation while increasing Strength, like evidence in Ryu \etal~\cite{GCFN}. Therefore, this justifies the fact that the ingredients are well-proposed to leverage low generalization error. We further support the theory through the analyses with the elements to showcase low validation errors in practice. We provide the following summary of our contributions:

\begin{enumerate}[label=(\roman*)]
\setlength{\topsep}{0pt}
\setlength{\itemsep}{0pt}
\setlength{\partopsep}{0pt}
\item We introduce a new network design principle to intensify a backbone by incorporating {\it multiple lightweight heads} instead of using a complicated head or expanding model in width and depth directions. 

\item We introduce a novel attention module that employs the Gramian of the penultimate features as a class token within an attention layer, thereby strengthening lightweight classifiers based on pairwise feature similarity. We call {\it Gramian attention}, which enhances expressiveness without compromising model speed.

\item We further propose a learning algorithm with a new loss that enforces multiple heads to yield less-correlated features to each other. Intriguingly, our learning method shows a faster convergence and yields strong precisions.

\item We provide an analysis tool for diagnosing design elements of a network and training methods to reveal the effectiveness of the proposed method based on Correlation and Strength with the generalization bound.

\end{enumerate}

\section{Method} 
This section first outlines the motivation for this work. We then present our network architecture and learning algorithm for less-correlated features.

\subsection{Motivation}
\noindent\textbf{Class tokens for class prediction.} Learning class tokens~\cite{vit, cait, xu2022multi, mmcap} have gained popularity because of their effectiveness and simplicity in training ViTs. The class tokens are fed right after the patchification layer for long interactions with features following the original design choice~\cite{vaswani2017attention}. Additionally, it has been revealed that having a shorter interaction on only later layers improves performance~\cite{cait}. Longer interactions may harm the discriminability of the class token due to the low-level features, while short interactions can effectively capture high-level information in the later class tokens. We adopt a short-interaction-like design for our network.

Second, using multiple class tokens~\cite{xu2022multi, mmcap} has contributed to enhancing the interactions' discriminability. They passively let the class token be learned upon a random initialization rather than actively using the class token. We notice that no studies have been conducted to strengthen the class token itself.
We argue that the way of utilizing class tokens in previous literature might not fully exploit the maximum capability of the learned model. We thus further imbue the class token by computing the Gramian from the feature to assign it as the class token.

\noindent\textbf{Employing multiple heads.} 
Previous works~\cite{InceptionResnet, yu2018deep_dla,sun2018fishnet,dft_eccv_2018,du2020spinenet,liu2018path,bi_fpn, mmcap} guide us that aggregating multiple features give significant benefits over single-path models such as ResNet~\cite{resnet}. Motivated by the success, we design our network learning multiple heads on the top of the backbone, barely spending a high computational budget; each head takes advantage of the aforementioned design manner. The design manner is also supported by the literature~\cite{random_forest, GCFN}, which tells us that weak learners (\ie, classifiers) should be strongly trained individually while diversifying the learned features for generalization. 

Furthermore, to learn stronger heads, we focus on the underexplored correlation among learned features~\cite{fpn,bi_fpn,du2020spinenet,lan2018knowledge,dft_eccv_2018}. We propose a so-called less-correlated learning method to maximize feature diversity. In this light, we believe designing a network that branches lightweight head classifiers instead of a complicated network is an appropriate option for making a good combination of the proposed learning method and architecture. 

\subsection{Our Network Architecture}%
\label{sub_sec:our_architecture}
\begin{figure}[t]%
\centering
\small
\begin{minipage}[b]{.3\textwidth}
\subfloat[\small{ViT~\cite{vit}}] {\label{subfig:token_vit}\includegraphics[width=0.93\textwidth]{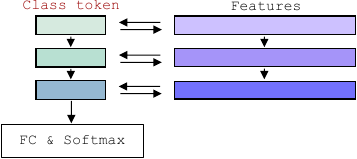}}
\vfill
\subfloat[\small{CaiT~\cite{cait}}] {\label{subfig:token_cait}\includegraphics[width=0.93\textwidth]{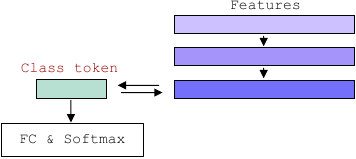}}
\end{minipage}
\hfill
\subfloat[\small{Ours}] {\label{subfig:token_ours}\includegraphics[width=0.15\textwidth]{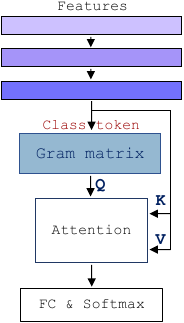}}
\vspace{-.5em}
\caption{ \small {\bf Contrast illustration of different heads}. We conceptually compare our head classifier design with class token-based classifier heads~\cite{vit,cait}. Ours employs Gramian with an attention layer to enhance the capability of the class token.}
\label{fig:token_comp}
\vspace{-.5em}
\end{figure}
\noindent\textbf{Gramian attention.}
We propose an attention-based module, dubbed Gram Attention, for aggregating visual tokens of a network more effectively. 
The primitive Transformer architecture with $n$-layers~\cite{vaswani2017attention, vit} uses the $C$-dimensional class token $Z \in \mathbb{R}^C$ to formalize the network output $Y$ as:
$Y=f^n\left(\dots f^1([Z; X])\right)$, where $[Z; X]$ denotes the concatenation of the $N$ visual tokens $X \in \mathbb{R}^{N \times C}$ and $Z$; $f^1$ and $f^{n}$ stand for the patch extractor and final classifier. This formulation indicates that an early concatenation of class tokens with the input-side features subsequently updates class tokens through long interactions with visual tokens (see Figure~\ref{subfig:token_vit}). Figure~\ref{subfig:token_cait} shows a variation~\cite{cait, mmcap} using the class token at later layers from an intermediate $m$-th layer, the output of which is formulated as $Y=f^n(\dots f^{m+1}\left([Z; f^{m}\left(...f^1\left(X\right)\right)]\right)$. Both of them are trained passively starting from the randomly initialized weights and may reach sub-optimal convergence points because optimizing both $X$ and $Z$ may not guarantee optimum at the same time. %
Unlikely, %
we alternatively assign the class token using features from a network. 

Figure~\ref{subfig:token_ours} illustrates our class token assigned by the Gramian computed with the penultimate features. This contrasts with the previous methods, where class tokens are initialized randomly and updated indirectly via feature interactions. 
Our aim is to directly influence entire trainable parameters in a network, so we compute the Gramian of the last-computed feature ${X}^{n-1}\in \mathbb{R}^{N \times HW \times C}$ at the penultimate layer $f^{n-1}$ as: 
\begin{equation}
Y=f^{n}\left([X^{n-1};  \mathcal{G}_{X^{n-1}}]\right), 
\label{eq:sahead}
\end{equation}
where $\mathcal{G}_X$ denotes the Gramian matrix of $X$ (\ie, an instance-wise computation $\mathcal{G}_X=X^TX \in \mathbb{R}^{N \times C \times C}$). We attribute the expressiveness of the Gramian to its computation of pairwise similarity. In practice, we compute $\mathcal{G}$ with the projected feature $V_{X} = X W_c$, where $W_c \in \mathbb{R}^{C \times \Tilde{C}}$. This is because the Gramian computation here has the complexity of $\mathcal{O}(HWC^2)$, and it becomes more computationally demanding with a large $C$, so we reduce it by $W_c$ to $\Tilde{C} \ll C$ for efficiency. Reducing the inner dimension also improves efficiency, but it could harm the encoded localization information. %
We introduce a Gramian computation with the vectorized feature to compute pairwise similarity across all locations by the following formula: 
\begin{align}
    \mathcal{G_X} = \text{Vec}(V_X^T V_X) W_g,
    \label{eq:gramian}
\end{align}
where $\text{Vec($\cdot$)}$ denotes the instance-wise vectorization, and $W_g \in \mathbb{R}^{\Tilde{C}^2 \times C}$ stands for another projection layer that restores the dimensionality to $C$, serving it as a class token for the subsequent attention layer. 

\noindent\textbf{Head classifier.}
Following Eq.~\eqref{eq:sahead}, $\mathcal{G}$ in Eq.~\eqref{eq:gramian} is fed into $f^n$ after concatenated with the input feature. We employ the attention~\cite{vaswani2017attention} as the final layer $f^n$. We refer to this layer, which computes the class embedding $Y$ as the head classifier. Note that the computed Gramian becomes the query, which is similar to \cite{cait}. Despite the shallow architecture, it has a large capacity standalone by the pairwise similarity computed by the Gramian. This associating operation is identical to the bilinear pooling~\cite{bilinear_pooling}, which has been revealed as learning strong spatial representation~\cite{mcb, bp_reid}. This operation is known to capture delicate spatial information across channel combinations, so it has been shown to improve the discriminative power of the object classification~\cite{bilinear_segmentation, bilinear_vqa, fine_grained_bilinear_compact}. We leverage the expressiveness of the bilinear representation for the class tokens possessing a strong spatial representation. 

\noindent\textbf{Extending to multi-head architectures.}
Constructing multiple branches on top of the backbone is a simple way to build multi-head classifiers.
We do not rely simply on the final feature but instead take the aggregated features from a backbone for the head classifiers.  
This is to take advantage of using diverse multi-level features similar to feature aggregation networks~\cite{fpn,yu2018deep_dla}.
Since we re-encode the aggregated features using lightweight heads, the multiple heads barely involve extra computational budgets. 
Therefore, our multi-head architecture can be regarded as an efficient alternative to heavy head architectures~\cite{sun2018fishnet,fpn,liu2018path,bi_fpn,aggre_du2020fine} or the way building complicated architectures ~\cite{aggre_zhao2021recurrence,efficientnet,zhai2022scaling}.%
\begin{figure*}[t]
\small
\hspace{-.2em}
\centering
\begin{subfigure}[t]{0.48\textwidth}
    \includegraphics[width=1.0\linewidth]{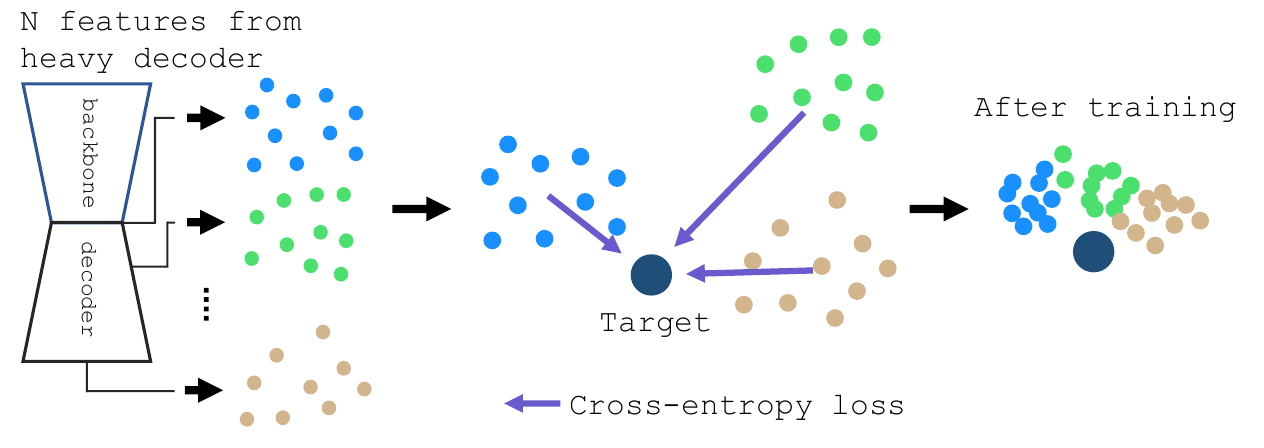}
    \caption{\small Traditional learning method}%
    \label{fig:workflow_base}
\end{subfigure}
\hfill
\begin{subfigure}[t]{0.48\textwidth}
\includegraphics[width=1.0\linewidth]{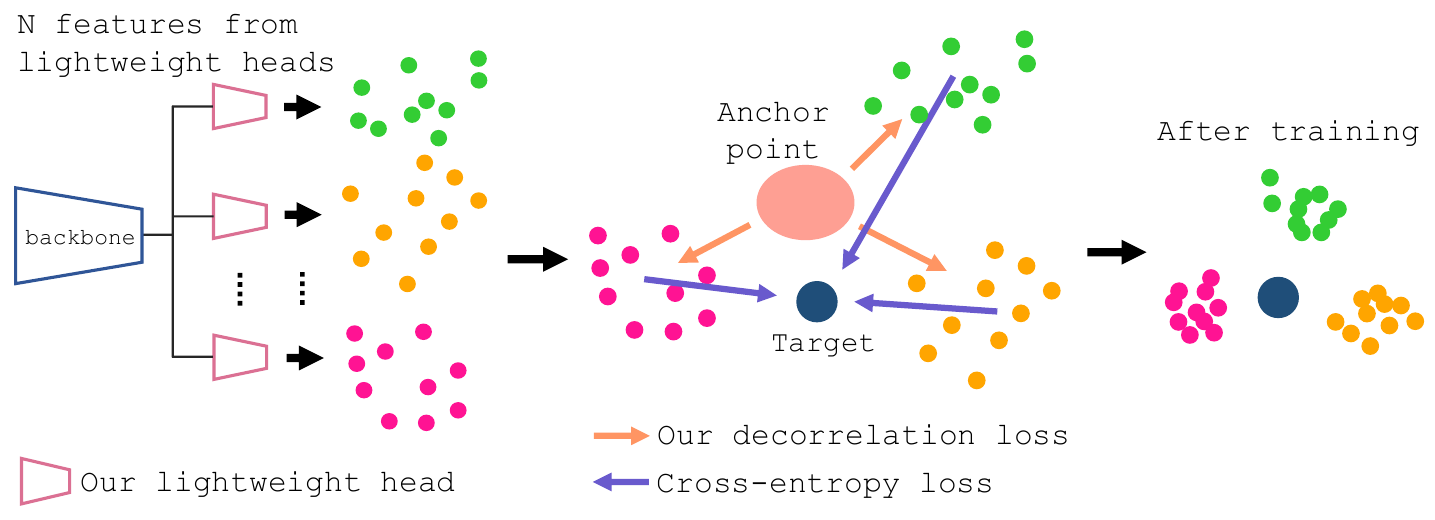}
    \caption{\small  Our learning method} %
    \label{fig:workflow_ours}
\end{subfigure} 
\vspace{-.5em}
\caption{\small {\bf Schematic illustrations.} We compare our learning method with a traditional one. (a) Multiple features extracted from different models (or a complicated head) are trained under the task loss only, so they are likely to get close to the ground-truth labels as training progresses; (b) our method trains lightweight heads by ensuring the representations are not highly correlated with each other.}
\label{fig:workflow}
\end{figure*}

\subsection{Training Multi-head Classifiers}

\noindent\textbf{On less-correlated multi-head classifiers.}
Here, we introduce a novel less-correlated learning method to learn more expressive multi-head classifiers. Training multiple identical network architectures or branches without considering feature diversity may not yield advantages. Since the models are expected to converge to nearby local minima during training, the resulting models are likely to learn correlated representations~\cite{Dropout,cogswell2015reducingdecorr, GCFN,huang2018decorrelated, hua2021featuredecorr} (see Figure~\ref{fig:workflow_base}). %
We begin with the model averaging loss (\ie, equally weighting the outputs) with the $i$-th output of $h$ heads as: 
\begin{equation}
\mathcal{L}=\sum_{i}^{h} {\mathtt{CE}_{i}} = -\sum_{i}^{h} y^T\cdot \log f^n_{i}(x),
\label{eq:vanilla_ens}
\end{equation}
where $\mathtt{CE}_i$ denotes the cross-entropy loss with the ground-truth label $y$, and $f_i(x)$ denotes the output of $i$-th head for the input $x$. For simplicity, we abbreviate $f^n$ (\ie, $n$-th layer's output) in previous notations to $f$.%

Directly minimizing Eq.~\eqref{eq:vanilla_ens}, the correlation among the predictions $f_i$ is likely to be high, so we propose a new decorrelation loss to avoid it: %
\begin{align}  
\label{eq:our_loss} 
&\mathcal{L}=-\sum_i^h y^T\cdot \log f_i(\hat{x}) + \lambda \mathcal{L}_{dec},\\
&\text{s.t.\ \ }\mathcal{L}_{dec}=\sum_i^h \sum_j\frac{f_j(\hat{x})^T}{n}\cdot \left(\log\sum_k\frac{f_k(\hat{x})}{n}-\log f_i(\hat{x})\right), \nonumber
\end{align}
where $\mathcal{L}_{dec}$ is coined by the decorrelation loss\footnote{We use the term {\it decorrelation} here in the idiomatic context of reducing the relevance and correlation of output predictions.} (see Figure~\ref{fig:workflow_ours}), and $\lambda$ is a tunable weighting parameter. Note that 
we use only negative $\lambda$ in Eq.~\eqref{eq:our_loss} to ensure the decorrelation loss functions in opposition to the cross-entropy loss. %

\noindent\textbf{Connection to knowledge distillation.}
One may speculate the proposed loss relates to Kullback Leibler Divergence used in the knowledge distillations~\cite{kd_hinton, fitnet}.
The canonical knowledge distillation methods use a positive value for $\lambda$; unlikely, our approach assigns a negative $\lambda$ in Eq.~\eqref{eq:our_loss}, which lets the knowledge from the aggregated prediction be reversely transferred. Therefore, each prediction $f_i$ would deviate and be less correlated (see Figure~\ref{fig:workflow_ours}). We argue that training with knowledge distillation (\ie, using the KD loss~\cite{kd_hinton}) may fail to let each head learn without high correlation. This result is evident that a positive $\lambda$ makes the distance between the aggregated prediction and each prediction get closer, so the predictions get similar, as shown in Figure~\ref{fig:workflow_base}. Our claim is addressed in the later discussion section, providing both qualitative and quantitive results (see the visualization in Figure~\ref{e} and compare it with Figure~\ref{d}. 

\begin{table}[t]
\small
\centering
\tabcolsep=0.25cm
\renewcommand{\arraystretch}{0.9}
\begin{tabular}{cccc|cc|c}
\toprule 
$\mathcal{C}$& dim & Gram & Dec & \begin{tabular}[c]{@{}c@{}}\footnotesize{FLOPs}\\\footnotesize{(G)}\end{tabular}  & \begin{tabular}[c]{@{}c@{}}\footnotesize{\#Params}\\\footnotesize{(M)}\end{tabular}  & \begin{tabular}[c]{@{}c@{}} Top-1\\ err (\%)\end{tabular} \\ 
\midrule
1 & 32 & - & - & 0.26 & 2.5 & 21.8 \\
1 & 32 & \checkmark & - & 0.26 & 2.6 & 21.2 \\
1 & 64 & \checkmark  & - & 0.35 & 4.0 & 19.8 \\
2 & 64 & \checkmark  & - & 0.26 & 2.6 & 20.1 \\
2 & 128 & \checkmark  & - & 0.35 & 4.1 & 18.4 \\
8 & 128 & \checkmark & - & 0.22 & 2.0 & 18.9 \\
\rowcolor{gray!10} 
8 & 128 & \checkmark & \checkmark & 0.22 & 2.0 & 16.9 \\ 
\bottomrule 
\end{tabular}
\vspace{-.5em}
\caption{\small {\bf Factor analysis}. The cardinality ($\mathcal{C}$) and the reduced input channel (dim) of the head classifiers are studied. We mainly verify the impact of the proposed Gramian attention (Gram) and decorrelation loss (Dec). We experiment with ResNet110 on CIFAR100. A careful design significantly improves accuracy without added computational costs.}
\label{tbl:abl_cifar}
\vspace{-.5em}
\end{table}

\begin{table}[]
\small
\centering
\tabcolsep=2mm
\renewcommand{\arraystretch}{0.9}
\begin{tabular}{c|c|ccc|c}
\toprule 
Net & Head & \#heads & $\lambda$ & {\footnotesize \#Params (M)} & {\footnotesize Top-1 acc (\%)} \\ 
\midrule
\multirow{5}{*}{\rotatebox[origin=c]{90}{ResNet50}} & GAP-FC & 1 / 10 &- & 25.9 / 44.0 & 75.3 / 75.7 \\
 & CaiT  & 1 / 5 & - & 21.8 / 38.5 & 76.7 / 77.0 \\
 & Gram  & 1 / 5 & 0 & 22.4 / 41.3 & \textbf{78.0} / 79.1 \\
 & Gram  & 1 / 5 & -0.4 & 22.4 / 41.3 & 77.9 / 79.2 \\
 & Gram  & 1 / 5 & -0.8 & 22.4 / 41.3 & 76.3 / \textbf{79.3} \\ \midrule
\multirow{6}{*}{\rotatebox[origin=c]{90}{ViT-S}} & GAP-FC & 1 / 20 & - & 22.1 / 29.4 &  76.3 / 76.3\\
 & ViT  & 1 / 20 & - & 22.1 / 29.4 & 75.3 / 75.4\\
 & CaiT & 1 / 5 & - & 22.8 / 27.3 & 75.2 / 75.3 \\
 & Gram & 1 / 5 & 0 & 22.9 / 27.7 & \textbf{78.3} / 78.4 \\
 & Gram & 1 / 5 & -0.4& 22.9 / 27.7 & \textbf{78.3} / 78.5 \\
 & Gram & 1 / 5 & -0.8& 22.9 / 27.7 & 78.2 / \textbf{78.9} \\ \bottomrule
\end{tabular}
\vspace{-.5em}
\caption{\small {\bf Extended factor analysis.} We extend the analysis to ImageNet-1K, building upon learned insights from Tab.~\ref{tbl:abl_cifar}. We study the impact of head types (Head), the number of heads (\#heads), and $\lambda$ in the decorrelation loss. We include the global average pooling with a fully-connected layer (GAP-FC), ViT, CaiT, and ours (Gram) shown in Fig.~\ref{fig:token_comp}. We report the accuracy of both single and multiple heads adjusted to have similar parameters (single/multiple heads).
}
\label{tbl:abl_in}
\vspace{-.5em}
\end{table}
\section{Experiment}
\label{sec:exp}
This section begins with the empirical analyses of the components of our method. We then demonstrate the superiority of our models through ImageNet classifications and transfer them to downstream tasks. 
We coin a network using our Gramian attention-included heads as \ours-network.

\subsection{Preliminary Factor Analyses} %
First, we study how each design element of the proposed method works on the CIFAR dataset. Table~\ref{tbl:abl_cifar} shows that using our proposed Gramian attention (Gram) and learning method with the decorrelation loss (Dec) boosts the accuracy significantly. The results also display a head classifier can be strengthened by increasing the aggregated dimension (dim) and cardinality ($\mathcal{C}$) under similar computational demands. Extending the analysis to the ImageNet-1K dataset, we investigate the effectiveness of our multiple head architectures and the proposed learning method in Table~\ref{tbl:abl_in}. All experiments are performed with identical network configurations to the baseline models (ResNet50, ViT-S), such as the stage configuration and channel dimension. 
We report accuracies training ResNet50s and ViTs for 50 and 100 epochs, respectively.

\begin{figure}[t]
\small
\centering
\hspace{-1em}
\subfloat[ResNet50] {\label{subfig:abl_lam_subnet_r50}\includegraphics[width=0.167\textwidth]{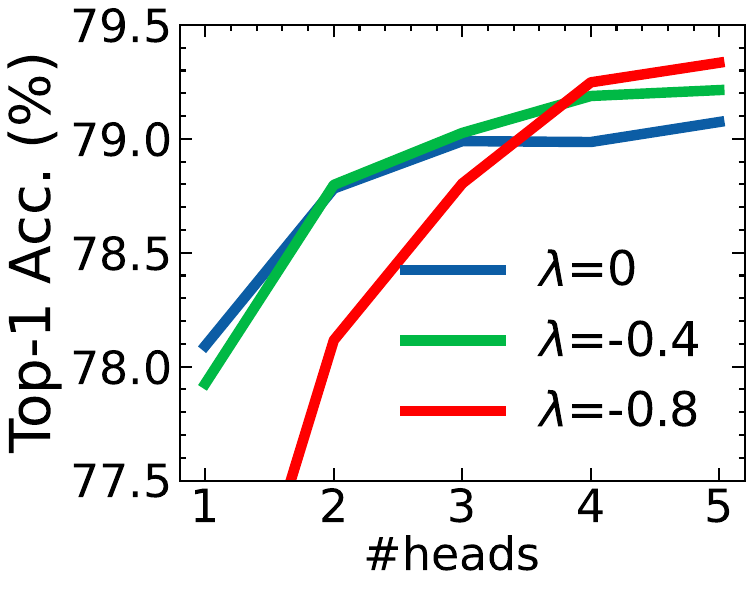}}
\subfloat[ConvNeXt-T]{\label{subfig:abl_lam_subnet_convnext}\includegraphics[width=0.16\textwidth]{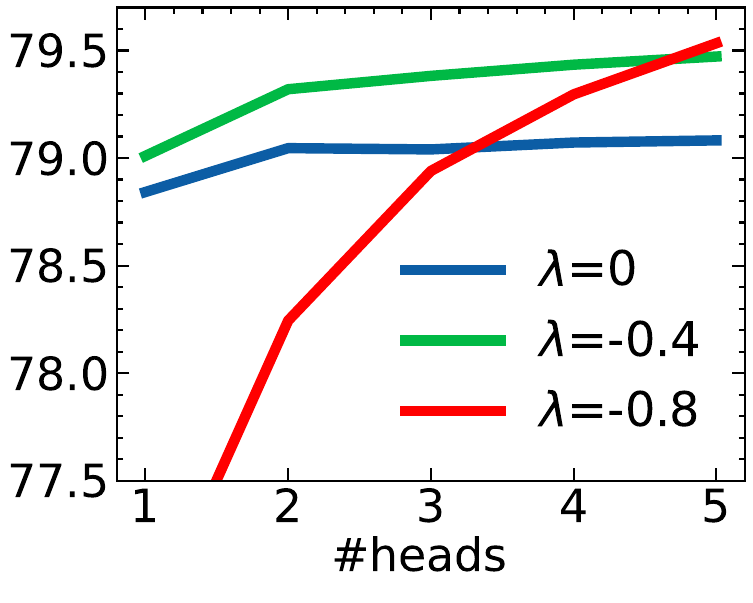}}
\subfloat[ViT-S] {\label{subfig:abl_lam_subnet_deit}\includegraphics[width=0.16\textwidth]{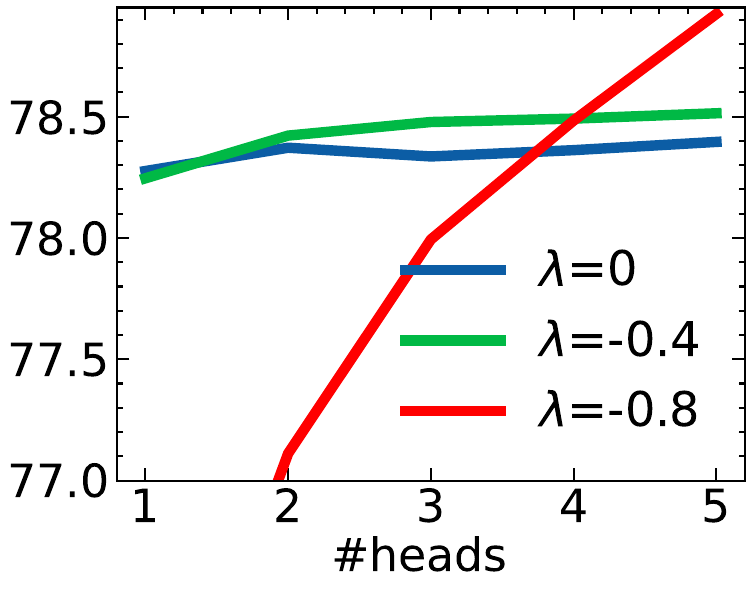}}
\vspace{-.5em}
\caption{\small
\textbf{Empirical study on \#heads and $\lambda$.} We examine ImageNet accuracy versus the number of head classifiers across different $\lambda$ for the decorrelation loss. Single-head underperforms, whereas using more heads increases the performance across all backbones. Lower values of $\lambda$ are more compatible with multiple heads; the best, with $\lambda{=}-0.8$, is achieved with five heads. 
} 
\label{fig:abl_wrt_lam_subnet}
\vspace{-.5em}
\end{figure}

\begin{table}[t]
\small
\tabcolsep=0.18cm
\begin{tabular}{l|rrc|c}
\toprule
Network         & \begin{tabular}[c]{@{}c@{}}\footnotesize{FLOPs}\\\footnotesize{(G)}\end{tabular}  & \begin{tabular}[c]{@{}c@{}}\footnotesize{\#Params}\\\footnotesize{(M)}\end{tabular} & \begin{tabular}[c]{@{}c@{}} \footnotesize{Throughput} \\ (img/sec)\end{tabular} &  \begin{tabular}[c]{@{}c@{}} \footnotesize{Top-1} \\ \footnotesize{acc} (\%)\end{tabular} \\ \midrule
\footnotesize{RSB-ResNet50~\cite{rsb}}     & 4.1  & 25.6  & 3409       & 79.8  \\
\rowcolor{gray!10} 
\ours-ResNet50   & 5.2  & 41.3  & 2145       & 82.5  \\ 
\footnotesize{RSB-ResNet152~\cite{rsb}}    & 11.6  & 60.2  & 1463       & 81.8  \\ \midrule
ViT-S~\cite{vit}          & 4.2 & 22.1  & 2556       & 79.8  \\
\rowcolor{gray!10} 
\ours-ViT-S     & 4.3   & 27.7  & 2289       & 80.9  \\
\rowcolor{gray!10} 
\ours-ViT-M     & 9.6  & 60.5  & 1322       & 82.6  \\
ViT-B~\cite{vit}          & 16.9 & 86.6  & 987        & 81.8  \\
\midrule
ConvNeXt-T~\cite{convnext}      & 4.5  & 28.6 & 2098       & 82.1  \\
\rowcolor{gray!10} 
\ours-ConvNeXt-T & 6.3  & 48.7 & 1452       & 83.2  \\
ConvNeXt-S~\cite{convnext}      & 8.7   & 50.2  & 1282       & 83.1  \\
\rowcolor{gray!10} 
\ours-ConvNeXt-S & 10.5 & 70.4 & 967        & 83.9  \\ConvNeXt-B~\cite{convnext}      & 15.4 & 88.6  & 903        & 83.8  \\
\rowcolor{gray!10} 
\ours-ConvNeXt-B & 19.0 & 124.3 & 668        & 84.3  \\ 
ConvNeXt-L~\cite{convnext}      & 34.4  & 197.8 & 507        & 84.3  \\
\bottomrule
\end{tabular}
\vspace{-.5em}
\caption{{\bf Our ImageNet-1K models.} We apply our method to the popular architectures, including ResNet~\cite{resnet}, ConvNeXt~\cite{convnext}, and ViT~\cite{vit,deit}; we dub our models \ours-ResNet, \ours-ConvNeXt, and \ours-ViT, respectively. %
All our models improve the baselines by large margins and enjoy faster speeds than each counterpart having similar accuracy. 
}
\label{tbl:scale_up}
\vspace{-.5em}
\end{table}

\begin{table}[!ht]
\centering
\small
\renewcommand{\arraystretch}{0.9}
\tabcolsep=0.18cm
\begin{tabular}{l|rrc|c}
\toprule
Network         & \begin{tabular}[c]{@{}c@{}}\footnotesize{FLOPs}\\\footnotesize{(G)}\end{tabular}  & \begin{tabular}[c]{@{}c@{}}\footnotesize{\#Params}\\\footnotesize{(M)}\end{tabular}  & \begin{tabular}[c]{@{}c@{}} \footnotesize{Throughput} \\ (img/sec)\end{tabular} &  \begin{tabular}[c]{@{}c@{}} \footnotesize{Top-1} \\ \footnotesize{acc} (\%)\end{tabular} \\ \toprule
\footnotesize{RSB-ResNet50~\cite{rsb}}     & 4.1  & 25.6  & 3409       & 79.8  \\
\footnotesize{RSB-ResNet152~\cite{rsb}}    & 11.6  & 60.2  & 1463       & 81.8  \\
ResNetY-8G~\cite{regnet}    & 8.0  & 39.2  & 827        & 82.1  \\
ViT-S~\cite{vit,deit} & 4.2  & 22.1  & 2556       & 79.8  \\
Swin-S~\cite{swin}          & 8.5  & 49.6 & 1024       & 83.0    \\
\footnotesize{PoolFormer-M36~\cite{poolformer}}  & 8.8   & 56.2   & 796        & 82.1  \\
CoatNet-0~\cite{coatnet}       & 4.2  & 27.4  & 1781       & 81.6  \\ 
CSwin-T~\cite{dong2022cswin}  & 4.3     & 23.0   & 1498 & 82.7 \\
ConvNeXt-S~\cite{convnext}      & 8.7   & 50.2  & 1282       & 83.1  \\
\midrule
\rowcolor{gray!10} 
\ours-ResNet50               & 5.2  & 41.3  & 2145       & 82.5  \\
\rowcolor{gray!10} 
\rowcolor{gray!10} 
\ours-ConvNeXt-T             & 6.3 & 48.7 & 1452       & 83.2
\\ \midrule
ResNetY-16G~\cite{regnet}       & 15.9 & 83.6 & 632        & 82.2  \\
ViT-B~\cite{vit,deit}    & 16.9 & 86.6 & 987        & 81.8  \\
Swin-B~\cite{swin}      & 15.1 & 87.8  & 731        & 83.5  \\
\footnotesize{PoolFormer-M48~\cite{poolformer}}   & 11.6 & 73.5  & 601        & 82.5 \\
CoatNet-1~\cite{coatnet}       & 7.6   & 41.7 & 985        & 83.3  \\ 
\footnotesize{InceptionNeXt-S~\cite{yu2023inceptionnext}} & 8.4  & 49 & - & 83.5 \\
CSwin-S~\cite{dong2022cswin}  & 6.9     & 35.0  & 933 & 83.6 \\ 
MaxViT-T~\cite{tu2022maxvit}   & 5.6  & 30.9 &  976  & 83.6  \\
SLaK-S~\cite{slak} & 9.8 & 55 & - & 83.8 \\
ConvNeXt-B~\cite{convnext}       & 15.4 & 88.6  & 903        & 83.8  \\ 
\midrule
\rowcolor{gray!10} 
\ours-CSwin-T & 6.1 & 42.0 & 1001 & 84.1 \\
\rowcolor{gray!10} 
\ours-ConvNeXt-S & 10.5 & 70.4 &967       & 83.9  \\ 
\midrule
ResNetY-32G~\cite{regnet}      & 32.3 & 145.1 & 378        & 82.4  \\
\footnotesize{InceptionNeXt-B~\cite{yu2023inceptionnext}} & 14.9 & 87  & - & 84.0 \\
SLaK-B~\cite{slak} & 17.1 & 95 & - & 84.0 \\
CoatNet-2~\cite{coatnet}       & 14.5  & 73.9 & 629        & 84.1  \\
CSwin-B~\cite{dong2022cswin}  & 15.0     & 78.0  &  549       & 84.2 \\ 
ConvNeXt-L~\cite{convnext}        & 34.4  & 197.8 & 507        & 84.3  \\
MaxViT-S~\cite{tu2022maxvit}  & 11.7  & 68.9  & 636  & 84.5  \\
CoatNet-3~\cite{coatnet}       & 32.5 & 165.2 & 360        & 84.5  \\ 
\midrule
\rowcolor{gray!10} 
\ours-ConvNeXt-B & 19.0 & 124.3 & 668 & 84.3  \\
\rowcolor{gray!10} 
\ours-ConvNeXt-B$^\dagger$ & 26.1  & 124.3  & 524     & 84.5 \\
\rowcolor{gray!10} 
\ours-CSwin-S & 8.7 & 54.3 & 671 & 84.7 \\ \bottomrule
\end{tabular}
\vspace{-.5em}
\caption{\small {\bf ImageNet-1K results.} Our models are compared with the state-of-the-art networks, including CNN, Transformer, and hybrid architectures on ImageNet-1K. We group the networks according to the computational budgets. All accuracies are borrowed from the original paper; RegNet accuracy is taken from ~\cite{rsb}. We report the throughputs measured by ourselves, running on an RTX 3090 GPU.
Our networks perform well over competitors with manageable resources and faster speed. We also provide the memory usage in the supplementary material.
$^\dagger$ uses $272 \times 272$ image size. \ours extremely improves CSwin family; we presume the lower channel dimension of CSwin architectures is an underlying reason.
} %
\label{tbl:SOTA_in} 
\vspace{-.5em}
\end{table}

As shown in Table~\ref{tbl:abl_in}, we confirm the models with proposed multiple heads significantly outperform baseline networks trained with the naive global average pooling (GAP-FC). %
Our Gramian attention remarkably outperforms existing ViT- and CaiT-like class token methods again. The proposed learning method with decorrelation loss (Dec) also contributes to performance, and this contribution is more significant with multiple heads and lowered $\lambda$ across all architectures. Figure~\ref{fig:abl_wrt_lam_subnet} gives more information on the accuracy variation of the models with multiple heads concerning $\lambda$ in the decorrelation loss. It verifies that the decorrelation loss can diversify learned features so that a higher $\lambda$ ($\lambda{=}-0.8$) performs better than other lower $\lambda$ cases ($\lambda{=}0$ and $\lambda{=}-0.4$).

\subsection{ImageNet Classification}
\label{sub_sec:img_classification}
\noindent\textbf{Implementation details.} 
We employ ResNet~\cite{resnet}, ConvNeXt~\cite{convnext}, CSwin~\cite{dong2022cswin}, and ViT~\cite{deit}  as our baseline networks, with each backbone branching out five heads. For ResNet50, we build our \ours-network with some popular tweaks; we reduce the channel dimension of the last three residual blocks to 1024 and exploit SE~\cite{SEnet}, and design tweaks introduced in the previous work~\cite{resnetd}. For ConvNeXt and ViT, we use the original architectures. For ViT, we encompass ViT-M having an intermediate model size between ViT-S and ViT-B, which has 576 channels with nine attention heads. For ConvNeXt and CSwin, due to the lower channel dimension compared to ResNet, we utilize a larger feature scale with minimal overhead.
Note that we do not delve into investigating more compatible backbones for our method architecturally. Instead, our focus is to showcase the effectiveness of our method through performance improvements on popular and straightforward network architectures under minimal resources.

\noindent\textbf{Comparison with state-of-the-arts.}
We compare the performance of \ours-networks with the contemporary state-of-the-art network architectures regarding the accuracy and computational complexities.
\ours-networks competes with the recently proposed network architectures, including the CNN architectures of RSB-ResNet~\cite{rsb}, RegNet~\cite{regnet}, ConvNeXt~\cite{convnext}, and SLaK~\cite{slak}; the ViT~\cite{vit}-related architectures, including ViT~\cite{deit}, Swin Transformer~\cite{swin}, and CSwin Transformer~\cite{dong2022cswin}; the hybrid architectures PoolFormer~\cite{poolformer}, CoatNet~\cite{coatnet}, MaxViT~\cite{tu2022maxvit}, and InceptionNeXt~\cite{yu2023inceptionnext}.
We systematically compare \ours-networks, including scaled-up models shown in Table~\ref{tbl:scale_up} with the competing models grouped by computational budgets, mainly focusing on throughput. Furthermore, we perform comprehensive comparisons with the popular contemporary models in Table~\ref{tbl:SOTA_in}, and it shows our models have clear advantages in throughput over their counterparts and outperform the competing networks, including the state-of-the-art CNN, ViT, and hybrid models. 

\begin{table}[t]
\small
\centering
\tabcolsep=0.25cm
\begin{tabular}{l|cc|c}
\toprule
Network  & AP (box) & AP (mask) & \#Params (M) \\ \midrule %
RegNetX-12G & 42.2  & 38.0 & 64.1           \\
Swin-T  &    42.7  & 39.3 &   47.8        \\ 
Poolformer-S36     &    41.0  &     37.7 &  31.6          \\ 
\midrule
\rowcolor{gray!10} 
\ours-R50 &  42.8 & 39.3   &  42.5     \\
\midrule
X101-64 & 48.3 & 41.7 & 140 \\ 
Swin-S & 51.9  & 45.0 & 107 \\
ConvNeXt-S & 51.9 & 45.0 & 108 \\
\midrule
\rowcolor{gray!10} 
\ours-ConvNeXt-S & 52.3 & 45.3 & 108 \\
\bottomrule
\end{tabular}
\vspace{-.5em}
\caption{\small \textbf{COCO instance segmentation results.} Our models ResNet50 (R50) and ConvNeXt-S outperform competing backbones using identical segmentation heads, respectively.}
\label{tbl:downstream_obj}
\vspace{-.5em}
\end{table}
\subsection{Downstream tasks}
We investigate the applicability of the proposed method to two downstream tasks, including instance segmentation and semantic segmentation.
Compared with the previous state-of-the-art models, we train pretrained \ours-networks on ImageNet-1k in Table~\ref{tbl:SOTA_in}.
Following the setups in literature~\cite{swin, poolformer, mmcap}, we attach detection and segmentation networks to ours. 
As in the literature~\cite{fpn, mmcap}, where the rear layers of the network are connected to the frontal layers, we attach dense prediction layers on our backbones.
We train our model with the widely-used MMDetection and MMSegmentation libraries\footnote{\scriptsize  \url{https://github.com/open-mmlab}}, and we report the performance of previous methods from the same training epochs or iterations.

\noindent\textbf{Object instance segmentation.}
We train the object instance segmentation model on COCO 2017~\cite{coco2017}. We exploit Mask R-CNN~\cite{maskrcnn} for ResNet-50 and Cascade Mask R-CNN~\cite{cascade_rcnn} for ConvNeXt-S as the baseline model. %
As shown in Table~\ref{tbl:downstream_obj}, ours outperform the models based on RegNet~\cite{regnet}, Swin Transformer~\cite{swin}, and PoolFormer~\cite{poolformer}.

\noindent\textbf{Semeantic segmentation.}
We train our models on the ADE20k semantic segmentation~\cite{ade20k}.
We employ two widely used heads: FPN~\cite{fpn} and UperNet~\cite{upernet} for the segmentation head in our model. As shown in Table~\ref{tbl:downstream_seg}, our networks exhibit competitive performance relative to models employing PoolFormer~\cite{poolformer} and Swin Transformer~\cite{swin} using each head.

\subsection{Training Setups}
\label{sub_sec:exp_setups}
\noindent\textbf{ImageNet-1K.}
Recent state-of-the-art networks~\cite{resnest, tresnet, rsb, xcit, tnt, deit} exploit training regimes with strong data augmentations, mostly based on timm library\footnote{\scriptsize \url{https://github.com/rwightman/pytorch-image-models/}}~\cite{timm}. We adopt a similar training regime, which employs Mixup~\cite{mixup}, CutMix~\cite{cutmix}, 
and RandAugment~\cite{cubuk2019randaugment} for data augmentation and use the cosine learning rate scheduling~\cite{loshchilov2016sgdr} with 300 epochs\footnote{In our ablation study in Table~\ref{tbl:abl_in}, we primarily train networks with ResNet-based and ConvNeXt-based models for 50 epochs and exceptionally train ViT-based models for 100 epochs due to its late convergence.
}. %

\begin{table}[]
\centering
\small
\tabcolsep=0.2cm
\begin{tabular}{l|l|cc|c}
\toprule
Head & Network  & Iter. & mIOU &   \#Params (M)  \\ %
\midrule
\multirow{2}{*}{FPN}     & PoolFormer-S36 & 40k &   41.6   &     34.6 \\ %
& \ours-R50        &  40k  &    \textbf{41.8}   &      26.6     \\ \midrule
\multirow{2}{*}{UperNet} & Swin-T   & 160k  &    44.4    &   59.9       \\
& \ours-R50        & 160k  &   \textbf{45.2}     &    67.3 \\ \bottomrule
\end{tabular}
\vspace{-.5em}
\caption{\small \textbf{ADE20k semantic segmentation results.} Our models outperform competing backbones with identical segmentation heads.}
\label{tbl:downstream_seg}
\end{table}

\noindent\textbf{Downstream task.}
For fair comparisons, we follow the same training setup of the competing backbones. We exploit $1 \times $ training schedule with 12 epochs in COCO. 
On ADE20k, we follow the same training setup of competitors again to train our segmentation model with iterations of 40k. 
We use 32 batch size for the 40k-iterations setup to compare ours with PoolFormer~\cite{poolformer} and use the 120k-iterations setup in Swin Transformer~\cite{swin} with 16 batch size.

\noindent\textbf{CIFAR100.}
We follow the standard 300-epochs training protocol with SGD~\cite{pyramidnet, cutmix} with the initial learning rate of $1\mathrm{e}{-3}$ decaying by $0.1$ at 150 and 225 epochs. We use 64 batch size for training using two GPUs. %

\section{Discussions}
\label{sub_sec:emp_study}
In this section, we investigate our method through the generalization bound analysis and the visualization method.

\begin{figure}[t]
\small
\captionsetup[subfloat]{labelformat=simple}
\centering
\subfloat{\label{subfig:abl_ge}\includegraphics[width=0.45\textwidth]{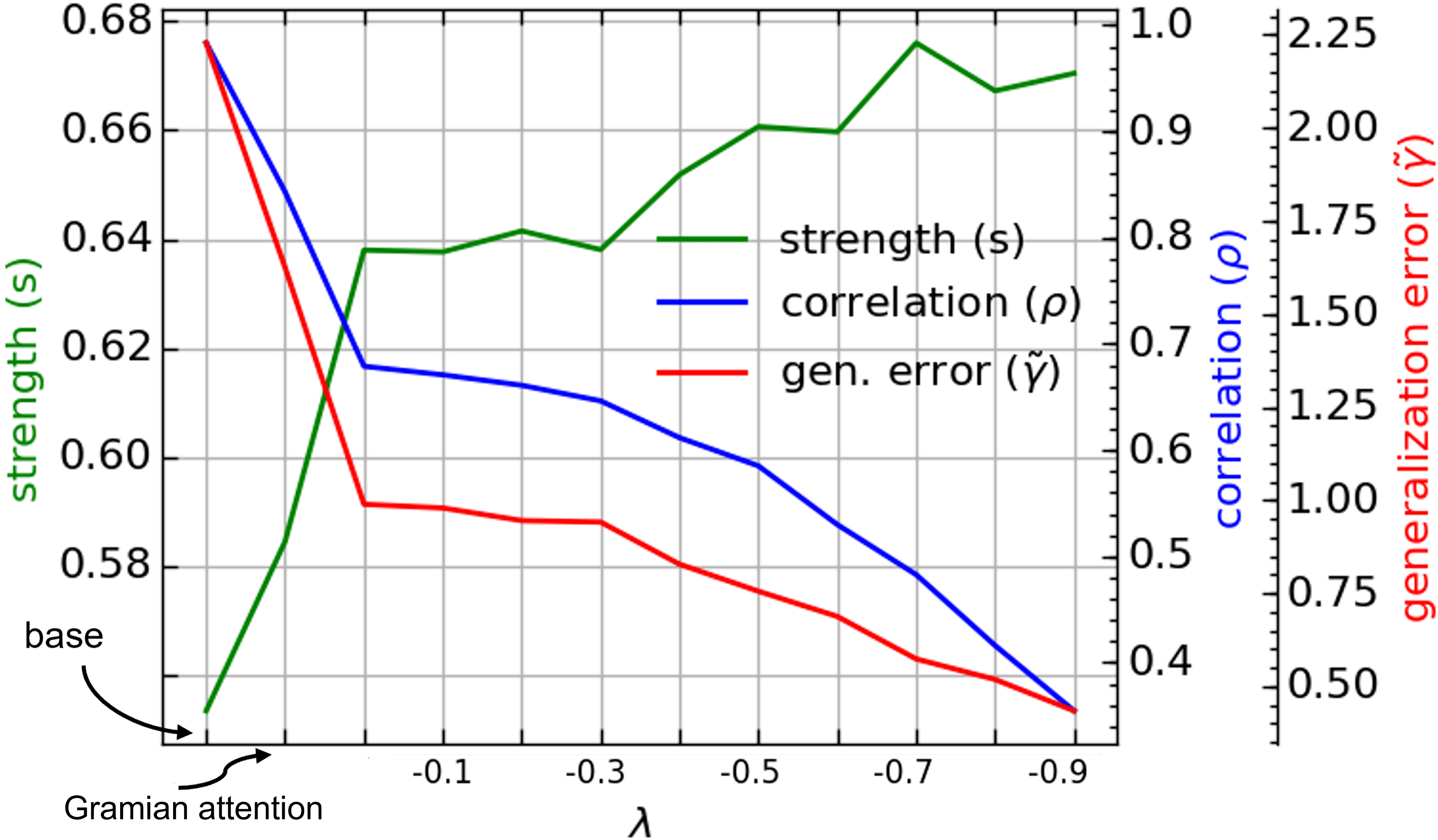}}
\vspace{-.5em}
\caption{\small {\bf Generalization error bound.} We visualize Correlation ($\rho$), Strength ($s$), and the upper bound of the generalization error ($\tilde{\gamma}$). We plot the metrics versus the architectural elements and different $\lambda$ values in the decorrelation loss. The left tick of the Gramian attention on the x-axis shows that architecture elements contribute to lowering the generalization error bound, and $\lambda$ in our less-correlated feature learning drops the bound on the right side. %
} %
\label{fig:generalization_bound}
\end{figure}

\subsection{Analyzing Our Method}
\label{sec:analysis}
Here, we justify our proposed design principle and learning method based on the foundation theory~\cite{random_forest, GCFN} that investigates the generalization capability of a model with multiple classifiers like ours. The theory is to compute the degree of Strength and Correlation for the generalization error bound~\cite{random_forest, GCFN}, and the magnitude of the metrics indicates how well the model generalizes~\cite{GCFN}.

\noindent\textbf{Strength and Correlation.}
Strength $s$ is firstly defined as the expectation of the margin between model prediction and the ground truth labels. The margin function is formulated as
$f({Y_{\phi}}, {\hat{Y}}) = P({Y_{\phi}}=\hat{{Y}})-\text{max}_{j\neq\hat{{Y}} } P({Y_{\phi}}=j)$,
where $Y_{\phi}$ and ${\hat{Y}}$ denote the output labels of a head classifier $\phi$ and the ground-truth labels of the data points, respectively. The last term $\text{max}_{j\neq\hat{{Y}} } P({Y_{\phi}}=j)$ stands for a set of labels with the largest probability amongst wrong answers. 

Correlation $\rho$ is computed with the raw margin function $\psi$, which is defined as %
$\psi({Y_{\phi}}, {\hat{Y}}) = \text{I}( {Y_{\phi}} = {\hat{Y}} ) - \text{I}( {Y_{\phi}} = \text{max}_{j\neq\hat{{Y}} } P({Y_{\phi}}=j))$,
where $\text{I}(\cdot)$ is the indicator function. $\rho$ is then computed by averaging the Pearson Correlation coefficient of $\psi$ between all combinations of heads $({\phi}_i, {\phi}_j)$.

\noindent\textbf{Generalization error bound.}
The upper bound of generalization error $\tilde{\gamma}$ is compute from Strength \(\mathcal{S}\) and Correlation \(\rho\), which is  %
$\gamma\leq\rho(1-s^2)/s^2$.
This implies Correlation and Strength are opposite to each other to achieve a low generalization error; however, importantly, the previous literature~\cite{GCFN} showed there could exist a method that trains a model to decrease Correlation while increasing Strength. Based on the evidence, we conjecture that an appropriate design of the head may also achieve it again. We confirm this by measuring the metric -- the generalization error bound -- to be reduced for particular architectural or training-related elements. Figure~\ref{fig:generalization_bound} shows that the proposed architectural design elements and learning method significantly reduce the upper bound of the generalization error. 

\begin{figure}[t]
\small
\centering
\hspace{-3mm}
\subfloat[Error vs. $\lambda$]{\label{subfig:abl_wrt_lambda}\includegraphics[width=0.25\textwidth]{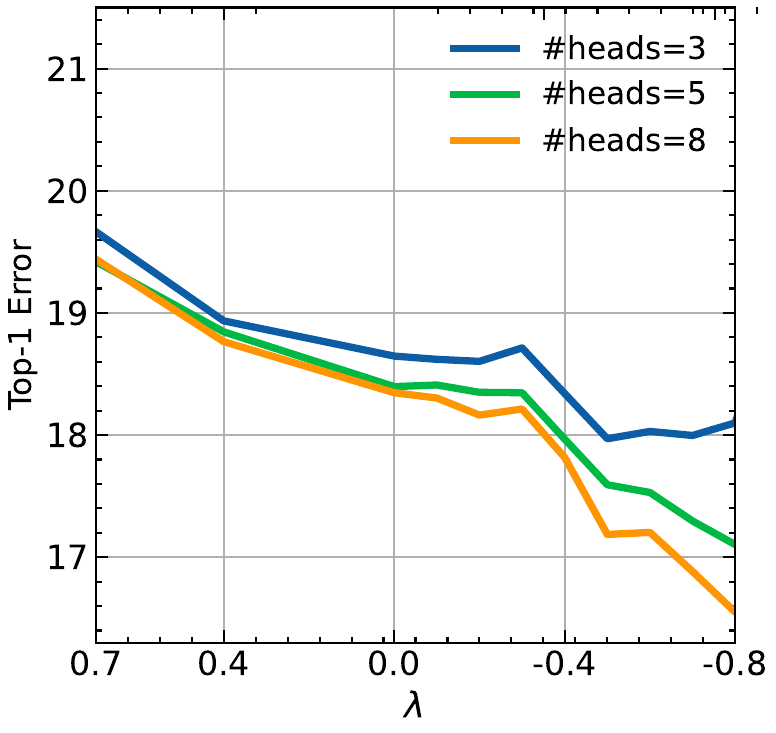}}
\hspace{-1mm}
\subfloat[Error vs. \#heads] {\label{subfig:abl_wrt_num_ltr}\includegraphics[width=0.24\textwidth]{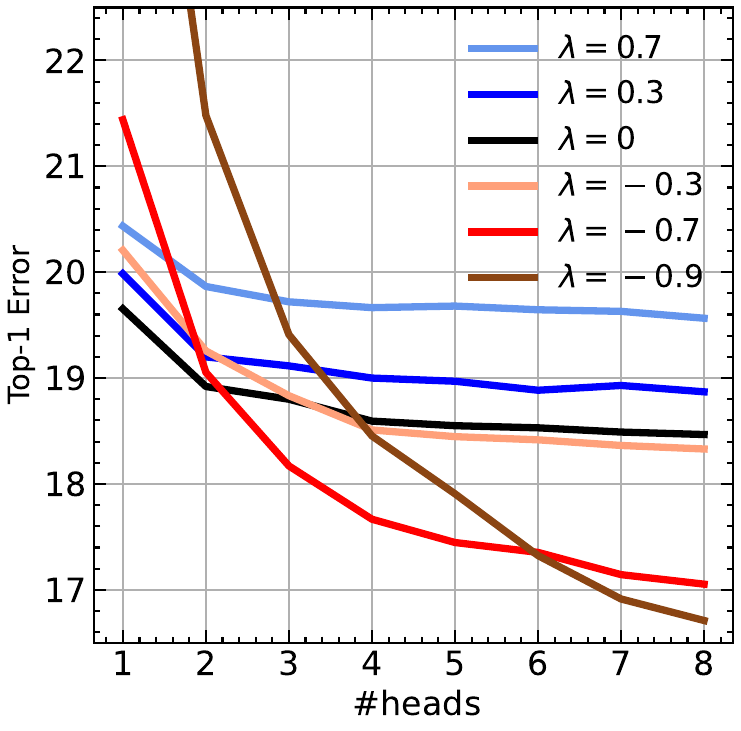}}
\vspace{-.5em}
\caption{\small {\bf Validation error trend w.r.t $\lambda$ and $\#$heads.}
(a) top-1 error versus $\lambda$ in the decorrelation loss; (b) top-1 error versus the number of heads. We observe that training with multiple heads with $\lambda<0$ significantly reduce the top-1 error.}
\label{fig:abl_wrt_dec}
\vspace{-.5em}
\end{figure}

We further visualize Correlation and Strength metrics together and observe Correlation gets consistently lowered as appending the architectural elements and adjusting the degree of the Correlation (adjusted by $\lambda$) in our learning method. This result indicates that our network architecture with multiple heads trained with our proposed learning method pushes the model to learn less-correlated and diversified features to contribute to the model's generalization capability. We further claim that the generalization bound is actually connected to performance in practice. We train models and visualize their validation errors in Figure~\ref{subfig:abl_wrt_lambda} and Figure~\ref{subfig:abl_wrt_num_ltr}. Along with Table~\ref{tbl:abl_in} reporting the error decreases as architecture advances, the figures show a consistent trend with Figure~\ref{fig:generalization_bound}.

\begin{figure*}[t]
\centering
\subfloat[\(\lambda=0\)]{\label{a}\includegraphics[width=0.155\textwidth]{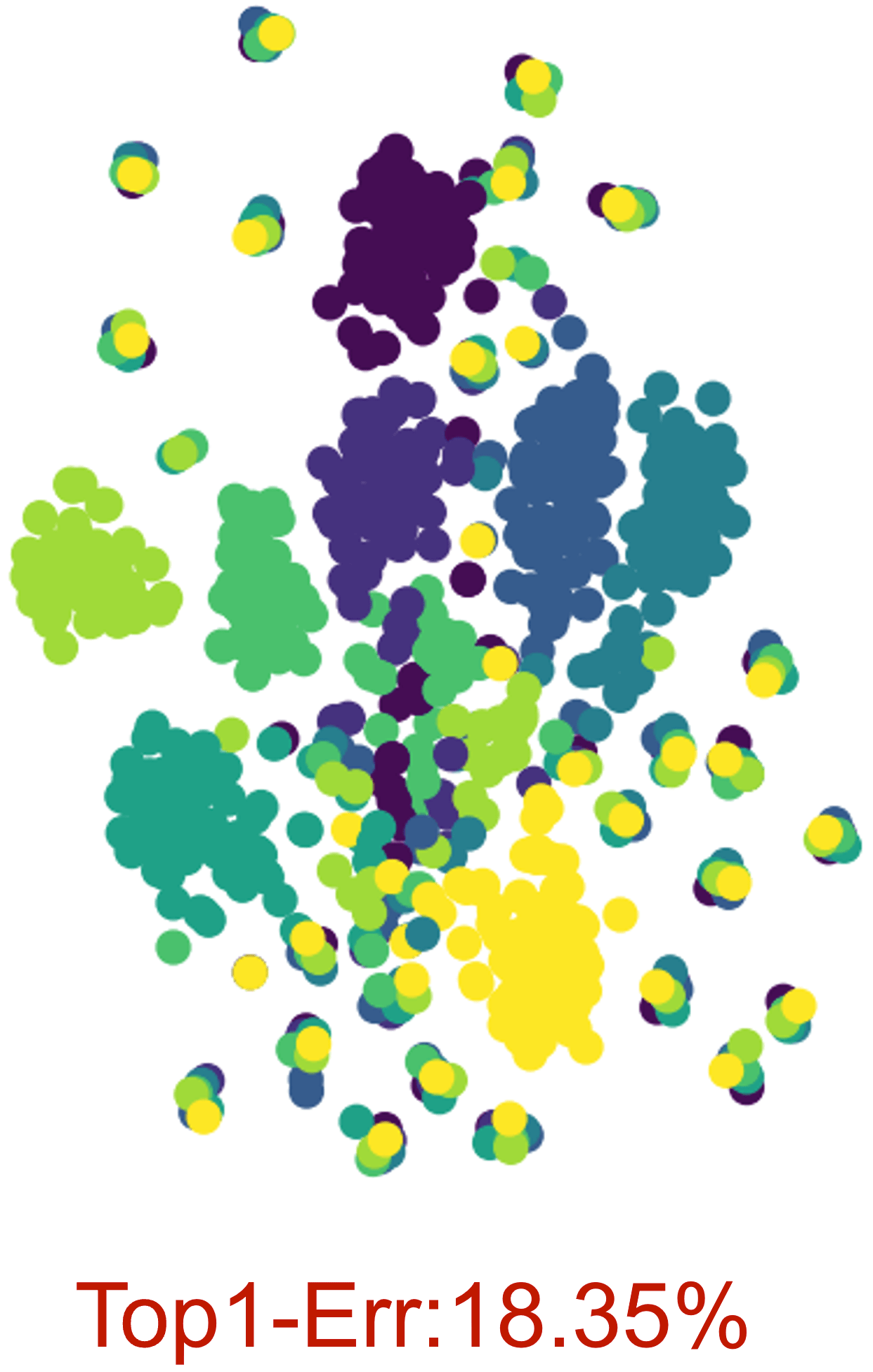}}
\quad\quad
\subfloat[\(\lambda=-0.3\)] {\label{b}\includegraphics[width=0.155\textwidth]{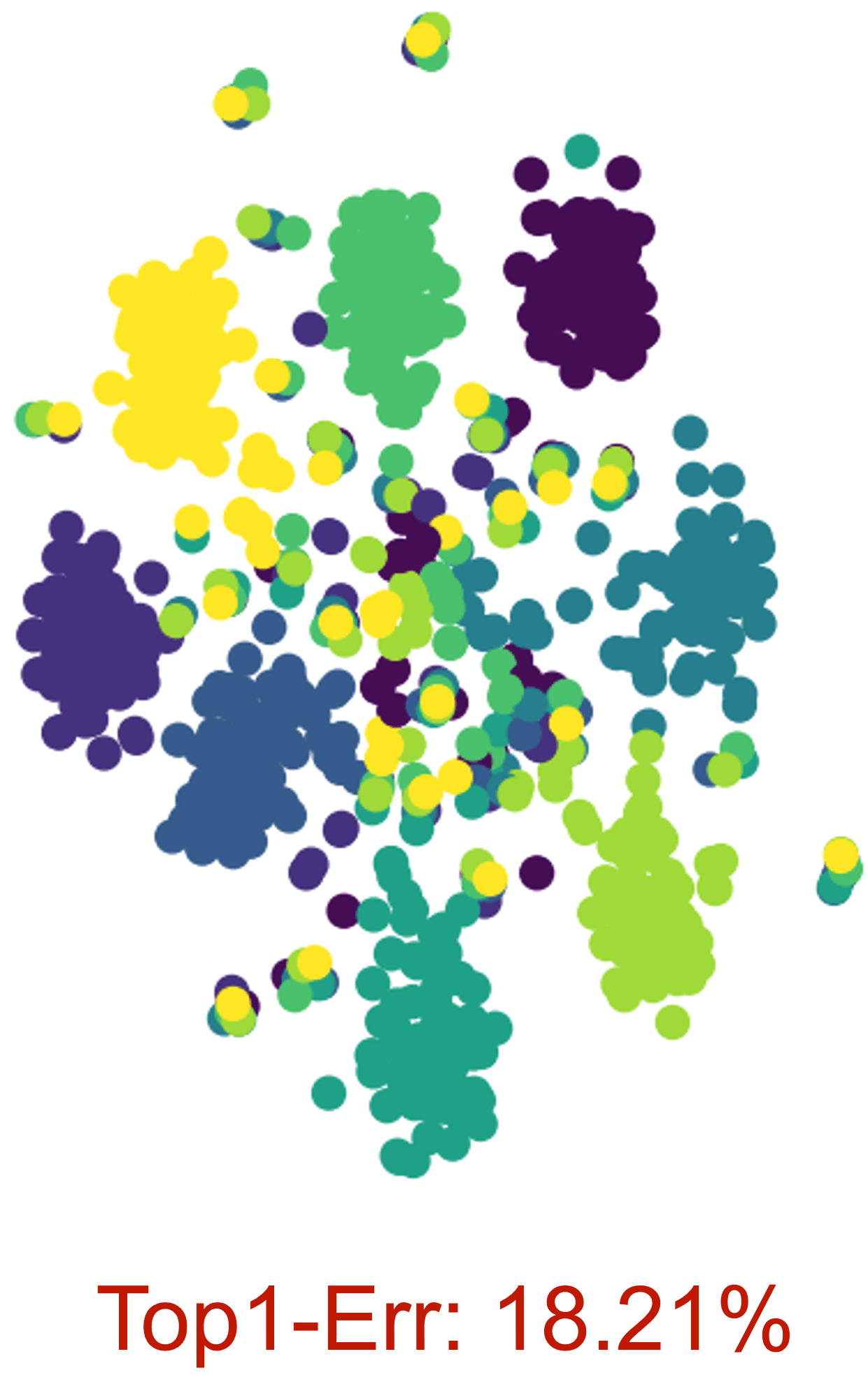}} \quad \quad
\subfloat[\(\lambda=-0.5\)] {\label{c}\includegraphics[width=0.155\textwidth]{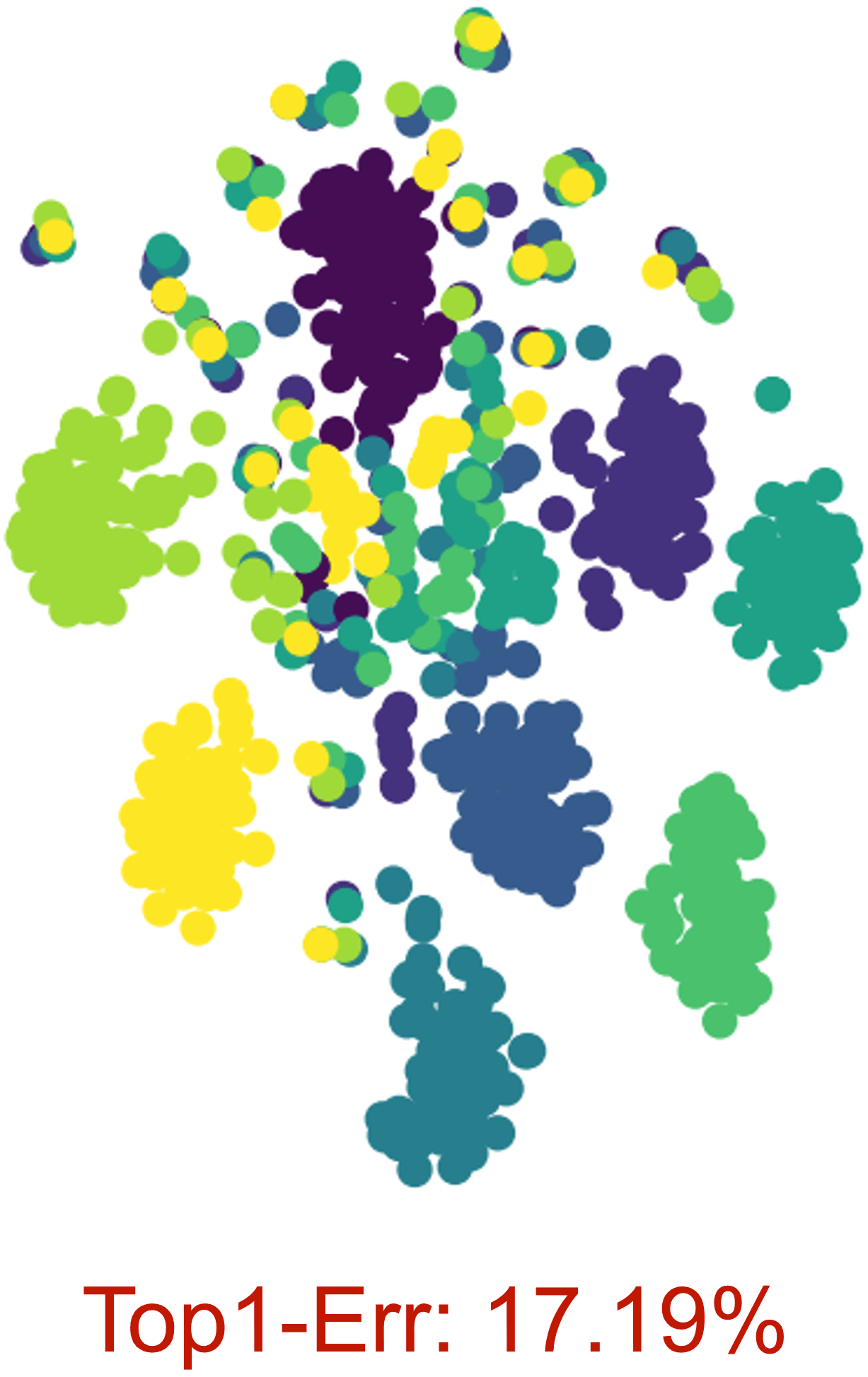}} \quad\quad
\subfloat[\(\lambda=-0.8\)] {\label{d}\includegraphics[width=0.155\textwidth]{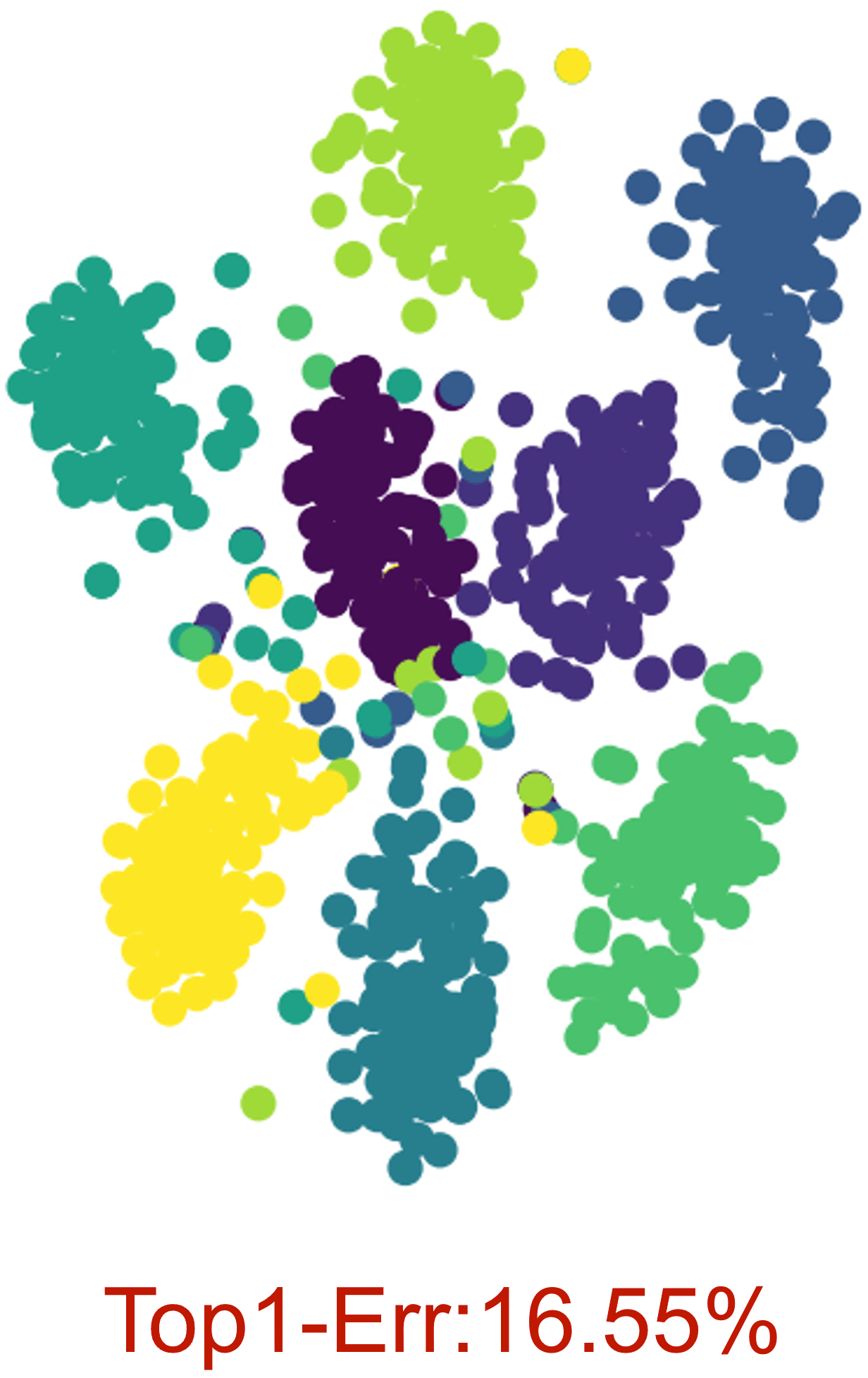}} \quad\quad
\quad \rulesep \quad \quad
\subfloat[\(\lambda=0.7\)]{\label{e}\includegraphics[width=0.155\textwidth]{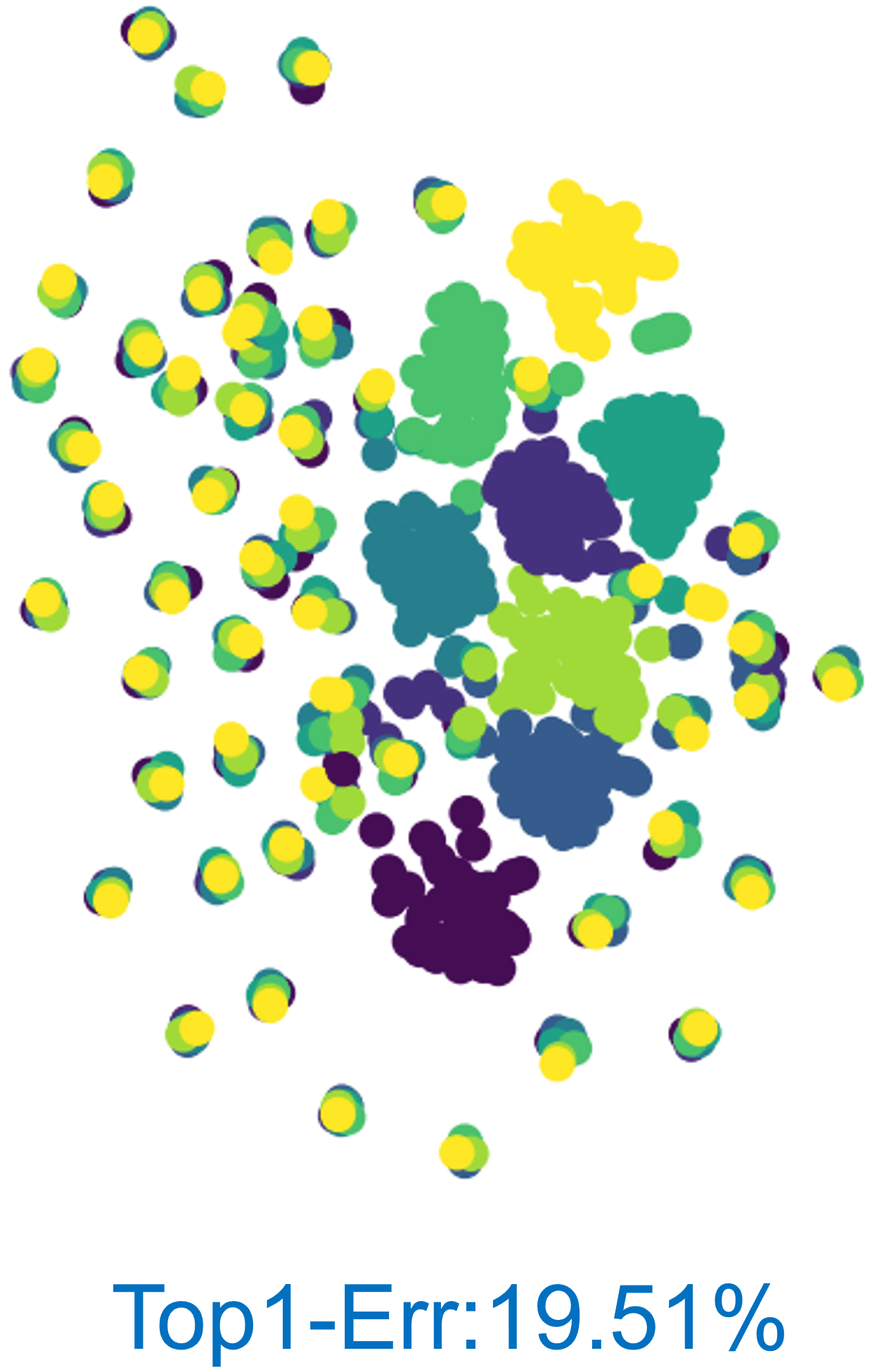}}
\vspace{-.5em}
\caption{\small {\bf t-SNE plots of the features extracted from learned head classifers}. We visualize how much the proposed learning method scatters the output features of each head. We extract the features from the images in the validation set and distinguish them from different head classifiers by color. We use features of a ResNet110 for (a) to (e). Specifically, (a) eight head classifiers without the decorrelation loss ($\lambda=0$); (b), (c), (d), and (e) different weighting parameters $\lambda$, respectively; We observe that 1) accuracy is aligned with the feature correlation; 2) our proposed learning method (\ie $\lambda<0$) works to increase the feature diversity with lowered correlation; 3) learned features with ($\lambda>0$) do not guarantee both low correlation and error (see (e).)}%
\label{fig:tsne_ltr_blocks}
\end{figure*}

\subsection{Visualizing Learned Features}
We investigate the impact of the decorrelation loss in Eq.~\eqref{eq:our_loss} with different  $\lambda$  by visualizing the output features with t-SNE~\cite{tsne}. Figure~\ref{fig:tsne_ltr_blocks} shows the clear trend when using $\lambda<0$; larger (to the negative direction) $\lambda$ let the model learn less-correlated features; the performance follows the trend. All with negative $\lambda$ outperforms the case of $\lambda=0$ that does not use the decorrelation loss. 

The performance with different $\lambda$ gets clearer with Figure~\ref{subfig:abl_wrt_lambda}, we achieve the best performance when the $\lambda$ is near -0.8, and when $\lambda>0$ the performance gets poorer than the model with $\lambda=0$. %
Additionally, a comprehensive visualization both with the number of heads and different $\lambda$ in Figure~\ref{subfig:abl_wrt_num_ltr} reveals some interesting aspects. We observe that when $\lambda$ reaches -0.7, the performance improves significantly as the number of heads increases. Performance gets saturated trained only with three heads when $\lambda\geq0$, while negative$\lambda$ lets the model avoid saturation.

\section{Related Work}
\noindent\textbf{Recent advance of the ImageNet networks.} 
After the emergence of ResNet~\cite{resnet}, EfficientNets~\cite{efficientnet} have dominated the field of ImageNet network architecture. Due to its low throughput compared to the low computational costs, ResNet~\cite{resnet} has been revisited by training it with more sophisticated training setups to maximize the performance and got new names called RS-ResNet~\cite{bello2021revisiting} and ResNet-RSB~\cite{rsb}. After the emergence of Vision Transformers (ViT)~\cite{vit}, DeiT~\cite{deit}, which trained ViT more effectively, invaded CNNs and got dominated. After that, another milestone was Swin Transformer~\cite{swin}, which pioneered the hierarchical ViT. A hybrid architecture such as CoatNet~\cite{coatnet} successively has showed another design principle using CNN and ViT effectively. ConvNeXt~\cite{convnext} was proposed to try to bring back the glory of CNN from ViT. Another hierarchical ViT, called CSwin~\cite{dong2022cswin}, showed more improved performance over Swin Transformer. Our work does not lie in a dominant trend of architectural development but is being studied to complement all architectures like a plug-and-play module.

\noindent\textbf{Network architectures with feature aggregation.}
Inception models~\cite{googlenet, bn, Inceptionv3, InceptionResnet} showed aggregating multiple features could further bring performance improvements. Veit \etal~\cite{resnet_like_shallow} interpreted ResNet~\cite{resnet} as an ensemble of numerous shallow neural networks, resulting in learning various features intrinsically. Inspired by \cite{newell2016stacked,fpn}, many previous works~\cite{sun2018fishnet,yu2018deep_dla,dft_eccv_2018,liu2018path,du2020spinenet,aggre_du2020fine, mmcap, aggre_zhao2021recurrence} proposed to design advanced architectures by aggregating multiple features. They heavily rely on multi-path connections with extra trainable layers as head architecture. Albeit they showed outstanding task performance, the models are computationally heavy due to additional learnable parameters; the multiple paths may learn similar representations. Our work shares a similar concept of aggregating features, but the difference is that we leverage a lightweight design regime for head classifiers instead of a complicated head architecture for a strong prediction through aggregation. Furthermore, it turns out that our lighter model consisting of the operations above achieves better discriminative powers with less correlated features.

\noindent\textbf{Training with lowering feature correlation.}
Despite the architectural advances, it has been reported that learned features are usually in high correlation~\cite{Dropout,cogswell2015reducingdecorr,huang2018decorrelated, zhang2018removingdecorr, GCFN,hua2021featuredecorr}. Algorithmic ways of training the features having a low correlation are also addressed in the literature~\cite{cogswell2015reducingdecorr,xiong2016regularizing,gu2018regularizing,zhu2018improvingdecorr}. Our method has, in a similar line to \cite{cogswell2015reducingdecorr,zhu2018improvingdecorr} which proposed distinctive losses that explicitly promote decorrelation at activation or filter, respectively. On the other hand, ours learn less-correlated features for aggregation in an inter-feature (or inter-layer) manner, directly affecting the final classifier. Lan \etal~\cite{lan2018knowledge} initially promoted ensemble branches by knowledge distillation, but the learned features were found to be highly correlated. Finally, it also turns out that our proposed architecture cooperates with the proposed learning technique towards improving the less-correlation property.

\section{Conclusion}
We have introduced a new learning framework with a network architecture leveraging lightweight heads. 
In contrast to traditional network architecture designs, we have proposed a novel approach using multiple lightweight head classifiers to create an expressive network. Our \ours-network aggregates the features refined by lightweight head classifiers, where the computational budget is significantly low. Additionally, our proposed learning method with the proposed decorrelation loss made our network learn less-correlated features, and aggregating them boosts performance due to learned complementary features. Our network has demonstrated increased feature diversification when employing the proposed learning method. The experimental results have proven that only the lightweight architecture has sufficient capacity for learning. We have analyzed our proposed method's effectiveness based on the Correlation and Strength theory. We found that the generalization bound has been consistently reduced for each proposed element and learning method. Finally, our network architecture has significantly outperformed the recent state-of-the-art CNNs, ViTs, and hybrid architectures on the ImageNet evaluation. Furthermore, several downstream tasks, including the COCO instance segmentation and ADE20k semantic segmentation, showcased our models' superior transferability. %
We expect our network design principle and method can be applied to any network architecture to improve performance. We hope the overall proposed framework facilitates future research.

\vspace{0.4em}
\noindent{\bf Limitations.} 
Even though the proposed design of employing lightweight multiple heads has minimal computational budgets, it unavoidably incurs extra parameters due to the internal channel dimension. We did not train extremely large baseline models such as large vision transformers such as ViT-H/14~\cite{vit} or ViT-G/14~\cite{zhai2022scaling}; we believe our method will be applicable to such large models.

\vspace{0.4em}
\noindent
\textbf{Acknowledgements.}
This work was supported in part by the Institute of Information and Communications Technology Planning and Evaluation (IITP) grant funded by the Korea Government (MSIT) (Artificial Intelligence Innovation Hub) under Grant 2021-0-02068 and under the Artificial Intelligence Convergence Innovation Human Resources Development (IITP-2023-No.RS-2023-00255968) Grant. 
{\small
\bibliographystyle{ieee_fullname}
\bibliography{gram_attention.bib}
}

\newcommand\blue[1]{{ #1}}

\iccvfinalcopy

\def\iccvPaperID{7292} %
\def\httilde{\mbox{\tt\raisebox{-.5ex}{\symbol{126}}}}

\ificcvfinal\pagestyle{empty}\fi

\renewcommand\thefigure{\Alph{figure}}
\setcounter{figure}{0}
\renewcommand\thetable{\Alph{table}}
\setcounter{table}{0}


\ificcvfinal\thispagestyle{empty}\fi

\section{Experiment (cont'd)}
\subsection{Fine-grained Visual Classification}
\paragraph{Training setup.}
To further investigate our pretrained models' transferability, we finetune the ImageNet-pretrained \ours-ResNet50 on the fine-grained visual classification (FGVC) datasets. We employ five datasets, including CUB-200~\cite{dataset_CUB}, Food-101~\cite{food101}, Stanford Cars~\cite{stanford_cars}, FGVC Aircraft~\cite{fgvc_aircraft}, and Oxford Flowers-102~\cite{flower102}. We grid-search the hyper-parameters similarly to \cite{kornblith2019do_imagenet, han2021rethinking} and follow the provided training regime for finetuning. We use the SGD optimizer with 20k iterations train networks and 224${\times}$224 center-cropped images from the downsized one to 256 on its shorter side.
We also finetune the ImageNet-pretrained ResNet50 on each dataset as baselines using identical grid-searches to exhibit maximal performance. Note that we report the accuracy at the final epochs rather than picking up the peak accuracy. We do not compare with the finetuning performance of other backbones due to the inherent differences in model size and finetuning training setup.
As shown in Table~\ref{tbl:transferlearning}, each of \ours-ResNet50s consistently outperforms their respective baseline counterparts.

\begin{table}[t]
\footnotesize
\centering
\tabcolsep=0.35em
\begin{tabular}{l|ccccc}
\toprule
Network & CUB-200 & Food & Cars & Aircraft & Flowers \\
\midrule
R50 & 79.2±1.7 & 86.9±0.2 & 90.4±1.0 & 86.2±2.0 & 99.0±0.4 \\
\midrule
\ours-R50 & \textbf{81.9±0.6} & \textbf{87.7±0.4} & \textbf{91.3±1.1} & 86.6±1.4 & 99.2±0.2 \\
\ours-R50 {\scriptsize (w/o heads)} & 81.4±1.0 & 87.1±0.1 & 90.7±1.0 & \textbf{86.8±1.4} & \textbf{99.3±0.2} \\
\bottomrule
\end{tabular}
\caption{\small {\bf Transfer learning results on FGVC datasets}. We compare the accuracy of the baseline ResNet50 and \ours-ResNet50 with or without heads. We report the averaged accuracies with the standard deviation to show the accuracy robustness across diverse hyper-parameter settings for each dataset. All the accuracies are reported by training and evaluation with 224${\times}$224 images. 
We observe that our approach demonstrates enhanced transferability. Surprisingly, even our model without heads (\ie, the backbone itself) exhibits improved transferability as well.}
\label{tbl:transferlearning}
\end{table}

\paragraph{Finetuining models without heads.}
We conjecture that our models would have empowered backbones (\ie, the models without head classifiers) having improved transferability. To validate this, we report the transfer learning performance of the finetuned backbone, which is identical to ResNet50 without the heads. Table~\ref{tbl:transferlearning} shows that \ours-ResNet50 (without head classifiers) enjoys consistent extra accuracy gains in Table~\ref{tbl:transferlearning}. We presume that our proposed method encourages the early layers (\ie, input-side layers) to learn more transferable representations due to the proposed lightweight heads that possess a few trainable parameters. We believe this shows a potential of utilizing our \ours-networks as a partial network without using heads at inference for further efficiency. We will give more results about employing partial networks in the later section.

\section{Additional Experimental Studies}
We conduct additional empirical studies with our proposed method. First, we showcase the capability of backbones that have no heads and models with only a single head by randomly removing all other heads. Second, we present comparative experiments with an existing multi-head neural network~\cite{lan2018knowledge}.

\begin{table}[t]
\small
\centering
\tabcolsep=0.7em
\begin{tabular}{l|l}
\toprule
Dataset & Accuracy gains (\%p) \\ \midrule %
CIFAR $(\text{depth=}29/65/110)$ & \blue{+0.49} / \blue{+0.85} / \blue{+0.24} \\ %
ImageNet / CUB / Food & \blue{+0.28} / \blue{+2.2} / \blue{+0.2} \\ 
Car / Aircraft / Flower & \blue{+0.3} / \blue{+0.6} / \blue{+0.3} \\ 
\bottomrule
\end{tabular}
\caption{{\bf Impact of our models without heads.} We study the backbone performance after eliminating heads. The numbers indicate top-1 accuracy gains over each baseline, which is trained with the identical setting to ours. This reveals our proposed method consistently improves the backbone's expressiveness across different datasets, detaching heads after training.
}
\label{tbl:abl_skd_cifar}
\end{table}

\subsection{Deploying Partial Networks}
\label{sub_sec:detach}
Our proposed method enables the deployment of partial networks from the overall learned network, enhancing efficiency (\ie, using the backbone alone or the network with fewer heads). In conjunction with Table~\ref{tbl:transferlearning}, the CIFAR and ImageNet results in Table~\ref{tbl:abl_skd_cifar} offer additional evidence that our backbones experience substantial improvement without heads, all without incurring extra computational demands. As aforementioned, using lightweight heads contribute to this improvement. We further speculate that this outcome arises due to the augmented gradients originating from multiple heads, which are learned through the proposed method. Furthermore, we argue that our decorrelation loss augments the gradients again, promoting less-correlatedheads. 

We adjust our models using only a single head classifier upon the baseline. We remove all the other heads but remaining a single head that is randomly chosen. The single head at the top of a backboneincurs minimal computational costs compared to the backbone itself yet achieves significant performance improvement, as shown in Table~\ref{tbl:ltr1_CIFAR}. In practice, the number of head classifiers can be adjusted to balance the accuracy, memory, and latency under resource limits.

\begin{table}[t]
\small
\centering
\tabcolsep=.5em
\begin{tabular}{c|c|cc|cc}
\toprule
Network & Depth & \begin{tabular}[c]{@{}c@{}}FLOPs\\ (G)\end{tabular} & \begin{tabular}[c]{@{}c@{}}\#Params\\ (M)\end{tabular}& \begin{tabular}[c]{@{}c@{}}Top-1\\ err (\%)\end{tabular} & \begin{tabular}[c]{@{}c@{}}Top-5\\ err (\%)\end{tabular} \\ \midrule
\multirow{3}{*}{R50} & 29 & 0.05 & 0.34 & 26.1 & 6.5 \\
& 65 & 0.10 & 0.71 & 22.0 & 4.9 \\
& 110 & 0.17 & 1.17 & 19.8 & 4.4 \\ \midrule
\multirow{3}{*}{\ours-R50} & 29 \hspace{0.15em} & 0.05 & 0.36 & \textbf{24.8} & \textbf{6.3} \\
& 65 \hspace{0.15em} & 0.11 & 0.76 & \textbf{21.2}&\textbf{4.8} \\
& 110 & 0.18 & 1.26 & \textbf{19.6} & \textbf{4.4}\\ 
\bottomrule
\end{tabular}
\caption{\small {\bf Impact of our models using only a single head.} We report a performance comparison of our models with a single head classifier with the ResNet baselines. The results show that only a single head classifier with negligible extra computational costs gives consistent and significant performance improvements.}
\label{tbl:ltr1_CIFAR}
\end{table}

\subsection{Comparison with Multiple Feature Learning}
Finally, we conduct additional experiments comparing with a prior multiple-feature learning method, which learns multiple features and aggregates. This is to show whether our method with lightweight heads actually works better than the method with heavy and complicated heads. Since such architectures~\cite{fpn,du2020spinenet} were aimed at different tasks, we choose a milestone work~\cite{lan2018knowledge} that also trains multiple high-level features from multiple branches for comparison. %

The branches in ONE-E~\cite{lan2018knowledge} appear similar to our head classifiers; however, ONE-E uses a copy of fractions in its backbone, resulting in overall heavy computational costs. Moreover, those branches are positioned differently compared with ours. ONE-E training highly relies on knowledge distillation to learn similar features among the branches, where the concept is completely distinct from ours. We argue that the reported improvements in the paper may stem from the heavy branches; they could learn expressive representations but are highly correlated to each other. 

To ensure a fair comparison, we employ the identical architecture proposed in the paper~\cite{lan2018knowledge} for training, which is found in the publicly released codebase\footnote{\url{https://github.com/lan1991xu/one_neurips2018}}, where there are three branches from the middle layer of ResNets. ResNet32 (R32) and ResNet110 (R110) are used for experiments, the standard network architectures for CIFAR training~\cite{pyramidnet, wideresnet}. We train the models for ONE-E and ours with identical training setups. Table~\ref{tbl:one_ens} shows that our models with the same number of head classifiers achieve better performance with extremely less computational demands.

\begin{table}[t]
\setlength{\tabcolsep}{0.1cm}
\small
\centering
\tabcolsep=.5em
\begin{tabular}{l|cc|cc}
\toprule
Method & \begin{tabular}[c]{@{}c@{}}FLOPs\\ (G) \end{tabular} & \begin{tabular}[c]{@{}c@{}}\#Params\\ (M) \end{tabular} & \begin{tabular}[c]{@{}c@{}}Top-1\\ err (\%)\end{tabular} & \begin{tabular}[c]{@{}c@{}}Top-5\\ err (\%)\end{tabular} \\ \midrule
ONE-E + R32 & 0.12 & 1.19 & 24.0 & 5.6\\
\ours~-R32 & {\bf 0.08} & {\bf 0.75} & {\bf 21.9} & {\bf 5.1}\\ \midrule
ONE-E + R110 & 0.29 & 2.96 & 19.9 & 4.3\\
\ours~-R110& {\bf 0.22} & {\bf 2.04} & {\bf 19.0} & {\bf 3.9} \\
\bottomrule
\end{tabular}
\caption{{\bf Comparison with the multiple feature learning method. }
We perform an experimental comparison of our method with ONE-E~\cite{lan2018knowledge}. Two baselines ResNet32 (R32) and ResNet110 (R110) are used, and ours consistently outperform the counterparts.}
\label{tbl:one_ens}
\end{table}

\subsection{Memory usage}
We measure the additional memory usage by our proposed method. As shown in Table \ref{tbl:mem}, the parameter overhead of our method is not severe, so the additional memory usage of them is manageable. 
Since this memory usage is mostly proportional to the parameters, other models with \ours will show similar trends.

\begin{table}[]
\small
\centering
\tabcolsep=0.75em
\begin{tabular}{lcccc}
\toprule
\multirow{2}{*}{Network} & \multirow{2}{*}{\#Params (M)} & \multicolumn{3}{c}{Memory} \\ \cmidrule{3-5}
 & & 128 & 256& 512 \\ \midrule
ViT-S& 22.1& 157.6& 231.1& 378.1\\
\ours-ViT-S & 27.7& 179.2& 252.7& 399.7 \\ \bottomrule
\end{tabular}
\caption{
{\bf Memory usage by batch size (\ie 128, 256, and 512).} We measure the memory usage of the input image tensor and parameters for ViT~\cite{vit,deit} models. 
}
\label{tbl:mem}
\end{table}

\end{document}